\documentclass[pdflatex,sn-mathphys]{sn-jnl}

\jyear{2021}%

\theoremstyle{thmstyleone}%
\theoremstyle{thmstyletwo}%

\theoremstyle{thmstylethree}%
\usepackage{float}
\usepackage{amsmath, bm}
\usepackage{threeparttable}
\usepackage{footnote}
\begin{document}

\title[Article Title]{CORE-PERIPHERY PRINCIPLE GUIDED REDESIGN OF SELF-ATTENTION IN TRANSFORMERS}


\author[1]{\fnm{Xiaowei} \sur{Yu}}
\equalcont{These authors contributed equally to this work.}

\author[1]{\fnm{Lu} \sur{Zhang}}
\equalcont{These authors contributed equally to this work.}

\author[2]{\fnm{Haixing} \sur{Dai}}
\equalcont{These authors contributed equally to this work.}

\author[1]{\fnm{Yanjun} \sur{Lyu}}

\author[2]{\fnm{Lin} \sur{Zhao}}

\author[2]{\fnm{Zihao} \sur{Wu}}

\author[3]{\fnm{David} \sur{Liu}}

\author*[2]{\fnm{Tianming} \sur{Liu}}
\email{tliu@cs.uga.edu}

\author*[1]{\fnm{Dajiang} \sur{Zhu}}\email{dajiang.zhu@uta.edu}

\affil[1]{\orgdiv{Department of Computer Science and Engineering}, \orgname{University of Texas at Arlington}, \orgaddress{\city{Arlington}, \postcode{76010}, \state{Texas}, \country{USA}}}

\affil[2]{\orgdiv{School of Computing}, \orgname{University of Georgia}, \orgaddress{\city{Athens}, \postcode{30602}, \state{Georgia}, \country{USA}}}

\affil[3]{\orgname{Athens Academy}, \orgaddress{\city{Athens}, \postcode{30602}, \state{Georgia}, \country{USA}}}

\abstract{Designing more efficient, reliable, and explainable neural network architectures is critical to studies that are based on artificial intelligence (AI) techniques. Numerous efforts have been devoted to exploring the best structures, or structural signatures, of well-performing artificial neural networks (ANN). Previous studies, by post-hoc analysis, have found that the best-performing ANNs surprisingly resemble biological neural networks (BNN), which indicates that ANNs and BNNs may share some common principles to achieve optimal performance in either machine learning or cognitive/behavior tasks. Inspired by this phenomenon, rather than relying on post-hoc schemes, we proactively instill organizational principles of BNNs to guide the redesign of ANNs. We leverage the Core-Periphery (CP) organization, which is widely found in human brain networks, to guide the information communication mechanism in the self-attention of vision transformer (ViT) and name this novel framework as CP-ViT. In CP-ViT, the attention operation between nodes (image patches) is defined by a sparse graph with a Core-Periphery structure (CP graph), where the core nodes are redesigned and reorganized to play an integrative role and serve as a center for other periphery nodes to exchange information. In addition, a novel patch redistribution strategy enables the core nodes to screen out task-irrelevant patches, allowing them to focus on patches that are most relevant to the task. We evaluated the proposed CP-ViT on multiple public datasets, including medical image datasets (INbreast) and natural image datasets (CIFAR-10, CIFAR-100, and TinyImageNet). Interestingly, by incorporating the BNN-derived principle (CP structure) into the redesign of ViT, our CP-ViT outperforms other state-of-the-art ANNs. In general, our work advances the state of the art in three aspects: 1) This work provides novel insights for brain-inspired AI: we can utilize the principles found in BNNs to guide and improve our ANN architecture design; 2) We show that there exist sweet spots of CP graphs that lead to CP-ViTs with significantly improved performance; and 3) The core nodes in CP-ViT correspond to task-related meaningful and important image patches, which can significantly enhance the interpretability of the trained deep model. (Code is ready for release).}




\maketitle

\section{Introduction}\label{sec1}
Aided by the rapid advancement in hardware and massively available data, deep learning models have witnessed an explosion of various artificial neural networks (ANN) architectures\cite{he2016deep, krizhevsky2017imagenet, vaswani2017attention}, and made breakthroughs in many application fields due to their powerful automatic feature extraction capabilities. It is widely expected the architectures of ANN, as the core of current AI techniques, to be more efficient, reliable, explainable, and transformable, to adapt to various and complex problems in real applications. Essentially, various ANN architectures, represented via different neuron wiring patterns, correspond to different information exchange mechanisms, and therefore, have an inevitable effect on the latent feature representation and the downstream task performance. For example, multilayer perceptron (MLP) directly stacks multiple layers of neurons with paired-wise full connections between adjacent layers, whereas convolutional neural networks (CNN) focus on learning effective convolutional kernels that indicate specific wiring patterns among the neurons within the receptive field. Similarly, recurrent neural networks (RNN) adopt cyclic connections between nodes, allowing output to affect subsequent input to the same nodes\cite{sherstinsky2020fundamentals}. This special neuron wiring pattern of building cycles between nodes also enables RNNs to model and infer temporal dynamic relationships\cite{tealab2018time} contained in sequential data. More recently, transformer has become another mainstream ANN architecture due to its outstanding self-attention mechanism that allows effective and efficient message exchanges among neurons, and produced promising results in the natural language processing\cite{vaswani2017attention, devlin2018bert} and computer vision domains\cite{dosovitskiy2020image, liu2021swin}. In particular, many advancements in transformer architecture design, e.g., vision transformer (ViT)\cite{dosovitskiy2020image}, have centered around more effective message exchange mechanisms among spatial tokens by designing different Token Mixers. For instance, the shifted window attention in Swin\cite{liu2021swin}, the token-mixing MLP in Mixer\cite{tolstikhin2021mlp}, and the pooling in MetaFormer\cite{yu2022metaformer}, among others, were all designed to improve the self-attention upon the original vanilla ViT\cite{dosovitskiy2020image}, and thus enable more effective and efficient message exchanges among spatial patches/tokens. However, despite tremendous advancements in ANN architecture design in MLPs, CNNs, RNNs, and transformers, particularly for better message exchange mechanisms, there has been a fundamental lack of general principles that can inform and guide such ANN architecture design and redesign.   

To seek such guiding principles for ANN architecture design, more and more research studies started exploring the “structural signatures" of well-performing ANNs. Hence, the deep learning community has witnessed a paradigm shift from optimal feature design to optimal ANN architecture design. In general, the major strategies for optimal ANN architecture design can be categorized into two basic streams based on how to search in the neural architecture space. The first strategy is to design neural architectures that achieve the best possible performance using given computing resources in an automated way with minimal human intervention. Neural architecture search (NAS)\cite{zoph2016neural, ren2021comprehensive, elsken2019neural} is a major methodology in this category. NAS has a relatively low demand for the researchers’ prior knowledge and experience, making it easier to perform modifications to the neural architecture though it usually comes with a high computational cost. The second category of the strategy is to take the advantage of prior knowledge from specific domains, such as brain science, to guide ANN architecture design. For example, the authors in \cite{zhang2021explainable} designed a two-stream model for grounding language learning in vision based on the brain science principle that humans learn language by grounding concepts in perception and action, and encoding “grounded semantics” for cognition. It is worth noting that the above-mentioned two strategies should be viewed as complementary to each other rather than being in conflict, and their combination provides the researchers with an opportunity to explore and design well-performing neural architectures under different principles. For instance, recent studies, via qualitatively post-hoc analysis, have found that the best-performing ANNs surprisingly resemble biological neural networks (BNN)\cite{you2020graph}, which indicates that ANNs and BNNs may share some common principles to achieve optimal performance in either machine learning or cognition/behavior tasks. 

\begin{figure}[!h]
\begin{center}
\includegraphics[scale=0.235]{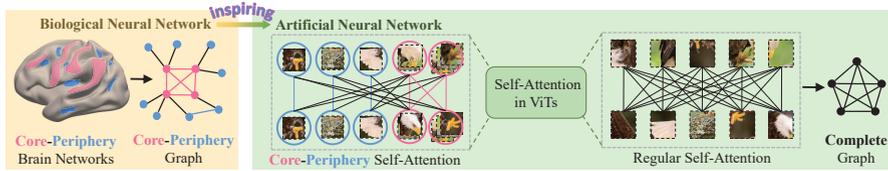}
\end{center}
\caption{The Core-Periphery principle in brain networks inspires the design of ANNs. The Core-Periphery structure broadly exists in brain networks, with a dense “core” of nodes (pink) densely interconnected with each other and a sparse “periphery” of nodes (blue) sparsely connected to the core and among each other. Inspired by this principle of BNN, we aim to instill the Core-Periphery structure into the self-attention mechanism and propose a new CP-ViT model.} 
\label{introduction}
\end{figure}

Inspired by the above-mentioned prior outstanding studies, in this work, we aim to proactively instill the Core-Periphery (CP) organization to guide the redesign of ANNs by using ViT as a working example. It has been widely confirmed that the Core-Periphery organization universally exists in the functional networks of human brains and other mammals, effectively promoting the efficiency of information transmission and communication for integrative processing\cite{bassett2013task, gu2020unifying}. The concept of the Core-Periphery brain network is illustrated in Fig. \ref{introduction}. By using the Core-Periphery property as a guiding principle, we infused its effective and efficient information communication mechanism into the redesign of ViT. To this end, we quantified the Core-Periphery property of the human brain network, infused the Core-Periphery property into ViT, and proposed a novel CP-ViT architecture. Specifically, we update the complete graph of dense connections in the original vanilla ViT\cite{dosovitskiy2020image} with a sparse graph with Core-Periphery property (CP graph), where the core nodes are redesigned and reorganized to play an integrative role and serve as a center for other periphery nodes to exchange information. Moreover, in our design, a novel learning mechanism is used to endow the core nodes with the power to capture the task-related meaningful and important image patches. We evaluated the proposed CP-ViT on multiple public datasets, including a medical image dataset (INbreast) and natural image datasets (CIFAR-10, CIFAR-100, TinyImageNet). The results indicate that the optimized CP-ViT in sweet spots\cite{you2020graph} outperforms other ViTs. We summarize our contributions in three aspects: 1) This work provides novel insights for brain-inspired AI: we can utilize the principles found in BNNs to guide and improve our ANN architecture design; 2) We show that there exist sweet spots of CP graphs that lead to CP-ViTs with significantly improved performance and 3) The core nodes in CP-ViT correspond to task-related meaningful and important image patches, which can significantly enhance the interpretability of the trained deep model.

\begin{figure}[h]
\begin{center}
\includegraphics[scale=0.083]{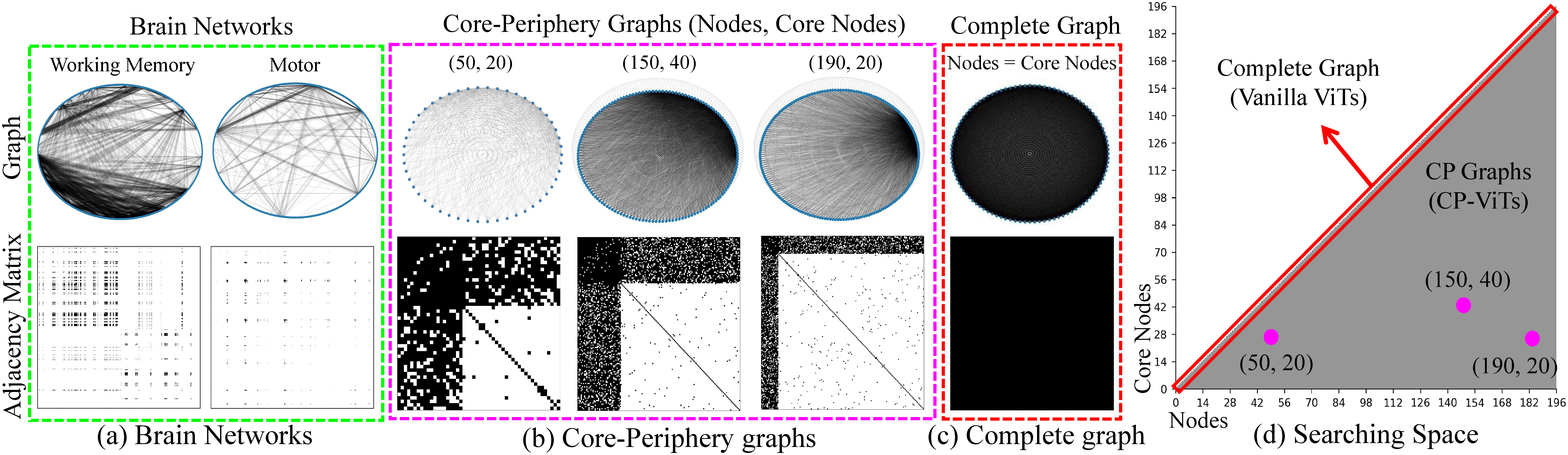}
\end{center}
\caption{ (a) Two types of representative brain networks in motor and working memory tasks. (b) Three examples of CP graphs. (c) Complete graph. The first row in (a), (b), and (c) shows their wiring patterns, while the second row shows their corresponding adjacency matrices. Black color in adjacency matrices means connections between nodes, while white represents no edge. (d) Graph search space defined by the total nodes number and the core nodes number. The complete graphs are located at the diagonal highlighted by a red box and the CP graphs are located at the remaining parts. }
\label{CPGraphs}
\end{figure}

\section{Results}\label{sec2}

\subsection{Exploring Core-Periphery Graphs}\label{subsec2}
\textbf{Core-Periphery property in brain networks.} We quantitatively measured the Core-Periphery property of brain networks. Working memory network (BN-WM) and motor network (BN-M) are two typical functional networks that are widely existed in the human brain. In this work, we used task fMRI data of these two tasks in the Human Connectome Project\cite{van2013wu} to generate functional brain networks. Using voxels as nodes and the correlations between fMRI signals associated with each voxel as edges, we built two population-level functional networks and showed their connection patterns as well as the adjacency matrices in Fig. \ref{CPGraphs}(a). To measure the Core-Periphery property of the two functional brain networks, we adopted independent probability\cite{cucuringu2016detection} as the measurement. Independent probability is defined as the probability that there is an edge between any pairs of nodes in a given matrix. Thus, the independent probabilities of the core-core connections, core-periphery connections, and periphery-periphery connections can be represented as  $I_{cc}$, $I_{cp}$ and $I_{pp}$, respectively. If the given matrix or graph is organized in a Core-Periphery manner\cite{holme2005core}\cite{rombach2014core}, the corresponding independent probabilities will have the following relations: $I_{cc} > I_{cp} > I_{pp}$. According to previous studies\cite{liu2019cerebral}, the convex gyri and concave sulci areas, which are two basic anatomical structures of the cerebral cortex, play different functional roles: gyri are functional hubs for global information exchange while sulci are responsible for local information processing. Therefore, we divided the nodes (voxels) into two categories, gyri-nodes (nodes in gyri regions) and sulci-nodes (nodes in sulci regions), and examined if brain networks have CP structure: gyri-nodes act as core nodes and sulci-nodes act as periphery nodes. The core-periphery measures of brain networks are shown in the last two columns in Table \ref{cp_value}. $R_{cc}$, $R_{pp}$ and $R_{cp}$ represent the normalized independent probabilities of core-core, core-periphery, and periphery-periphery connections. The independent probabilities and normalized independent probabilities are formulated as:
\begin{equation}\label{normalizedIP}
\begin{aligned}
I_{cc} = \frac{  1_{A_{cc}} }{ \lVert{A_{cc}} \rVert_{1} },
I_{cp} = \frac{  1_{A_{cp}} }{ \lVert{A_{cp}} \rVert_{1} },
I_{pp} = \frac{  1_{A_{pp}} }{ \lVert{A_{pp}} \rVert_{1} },\\
R_{cc} = I_{cc} / (I_{cc}+ I_{cp} + I_{pp}),\\
R_{cp} = I_{cp} / (I_{cc}+ I_{cp} + I_{pp}),\\
R_{pp} = I_{pp} / (I_{cc}+ I_{cp} + I_{pp}).
\end{aligned}
\end{equation}

\noindent \textbf{Core-Periphery structure in artificial neural networks.} We introduced the Core-Periphery organization into ANNs by CP graphs. There are two key factors that can affect the CP graph generation process. The first is the number of nodes, including the number of total nodes and the core nodes, which defines the search space. In this work, we set the maximum number of total nodes as 196, i.e., the number of patches for the vision transformer, then the number of core nodes can be any number between 0 and 196. Thus, the search space will include $\sum_{i=1}^{196}\sum_{j}^{0<j<=i} (i+j) = 19208 $ types of CP graphs, where $i$ and $j$ represent the number of total nodes and the core nodes.  The second is the wiring patterns of CP graphs: in this work, we used $p_{cc}$, $p_{cp}$, and $p_{pp}$ to represent the wiring probabilities between core-core nodes, core-periphery nodes, and periphery-periphery nodes, respectively. Fig.\ref{CPGraphs} (b) and (c) present the wiring patterns and adjacency matrices of three examples of CP graphs and the complete graph. As shown in Fig. \ref{CPGraphs}(b) and (c), CP graphs are densely connected for core nodes and sparsely connected for periphery nodes. The overall connection patterns of CP graphs are more sparse than the complete graph. The search space of CP graphs was shown in Fig. \ref{CPGraphs}(d) where the complete graphs located at the diagonal were highlighted by a red box and three types of CP graphs corresponding to Fig. \ref{CPGraphs}(b) were highlighted by pink circles. For each type of CP graph, we generated 5 samples with different wiring patterns and obtained 19208 * 5 CP graphs in total. Since the number of the generated CP graphs is huge (19208 * 5 in total), we sampled 190 types of CP graphs out of the total 19208 and finally obtained 190*5 candidates. For example, for a CP graph with 50 nodes, the number of core nodes is set to be [10, 20, 30, 40]. As a result, four different CP graphs, including [50, 10], [50, 20], [50, 30], and [50, 40], are obtained. For each of these four types of CP graphs, we generate 5 samples for further experiments.

\begin{table}[t]
\caption{ Evaluation of the Core-Periphery property in CP graphs, graphs generated by other graph generators, and brain networks} 
\centering
\begin{tabular}{ccccccc}
\hline
\multirow{1}{*}{IP}  
                   & CP Graphs  & CE. Graphs  & WS Graphs  & ER Graphs & BN-M & BN-WM     \\
\hline
$R_{cc}$  &$.59\pm{.06}$    & $.33\pm{.00}$    &  $.40\pm{.27}$  &$.36\pm{.23}$ & $.55\pm{.11}$ & $.61\pm{.09}$\\
$R_{cp}$   &$.35\pm{.13}$    & $.33\pm{.00}$    &$.40\pm{.28}$    &  $.36\pm{.24}$ & $.34\pm{.07}$ & $.26\pm{.10}$ \\
$R_{pp}$  &$.07\pm{.06}$   &  $.33\pm{.00}$    & $.20\pm{.28}$    &  $.28\pm{.22}$ & $.15\pm{.05}$ & $.14\pm{.06}$ \\
\hline
\label{cp_value}
\end{tabular}
\end{table}

Similar to brain networks, we also used the normalized independent probability to measure the Core-Periphery property for the generated CP graphs. We calculated the normalized averaged independent probability over 190*5 CP graphs and showed the results in the first column of Table \ref{cp_value}. From the table we can see that $R_{cc} > R_{cp} > R_{pp}$, which suggests that our generated CP graphs, as expected, display prominent Core-Periphery properties, while the graphs generated by the classic graph generators, such as  (1) Complete graph (CE.) generator; (2) Watts-Strogatz (WS) generator; and (3) Erdos-Renyi (ER) generator don't have the Core-Periphery property.

\begin{table}{H}
\caption{Summary of datasets}
\centering
\begin{tabular}{cccccc}
\hline
\multirow{1}{*} Dataset & Training &  Validation  & Class  & Original Res. & Resized Res.\\
\hline
INbreast & 6000  & 100   & 3   & 1024 * 1024 * 3  & 224 * 224 * 3   \\
CIFAR-10 & 50000    & 10000  & 10 & 32 * 32 * 3 & 224 * 224 * 3\\
CIFAR-100 & 50000   & 10000  &  100  & 32 * 32 * 3 & 224 * 224 * 3\\
TinyImageNet & 100K  & 10000  & 200 & 64 * 64 * 3 & 224 * 224 * 3\\
\hline
\label{Datasets}
\end{tabular}
\end{table}

\subsection{Sweet Spots for CP-ViTs}
In this section, we evaluated the performance of the proposed CP-ViT. The CP-ViT was implemented based on the ViT-S/16 architecture\cite{chen2021vision} and evaluated on $4$ different types of public datasets, the medical image dataset INbreast\cite{moreira2012inbreast}, the natural image dataset CIFAR-10\cite{krizhevsky2009learning}, CIFAR-100\cite{krizhevsky2009learning} and TinyImageNet\cite{griffin2007caltech}. The summary of the datasets we used in this work is presented in Table \ref{Datasets}. The parameters of CP-ViT were initialized and fine-tuned from ViT-S/16 trained on ImageNet\cite{krizhevsky2017imagenet}. We trained the CP-ViT for 100 epochs with batch size $64$ for INBreast and $256$ for CIFAR-10, CIFAR-100 and TinyImageNet, and used AdamW optimizer and cosine learning rate schedule\cite{loshchilov2016sgdr} with an initial learning rate of $0.0001$ and minimum of $1\mathrm{e}{-6}$. All the experiments were conducted using NVIDIA Tesla V100 GPU. 

\begin{figure}
\begin{center}
\includegraphics[scale=0.130]{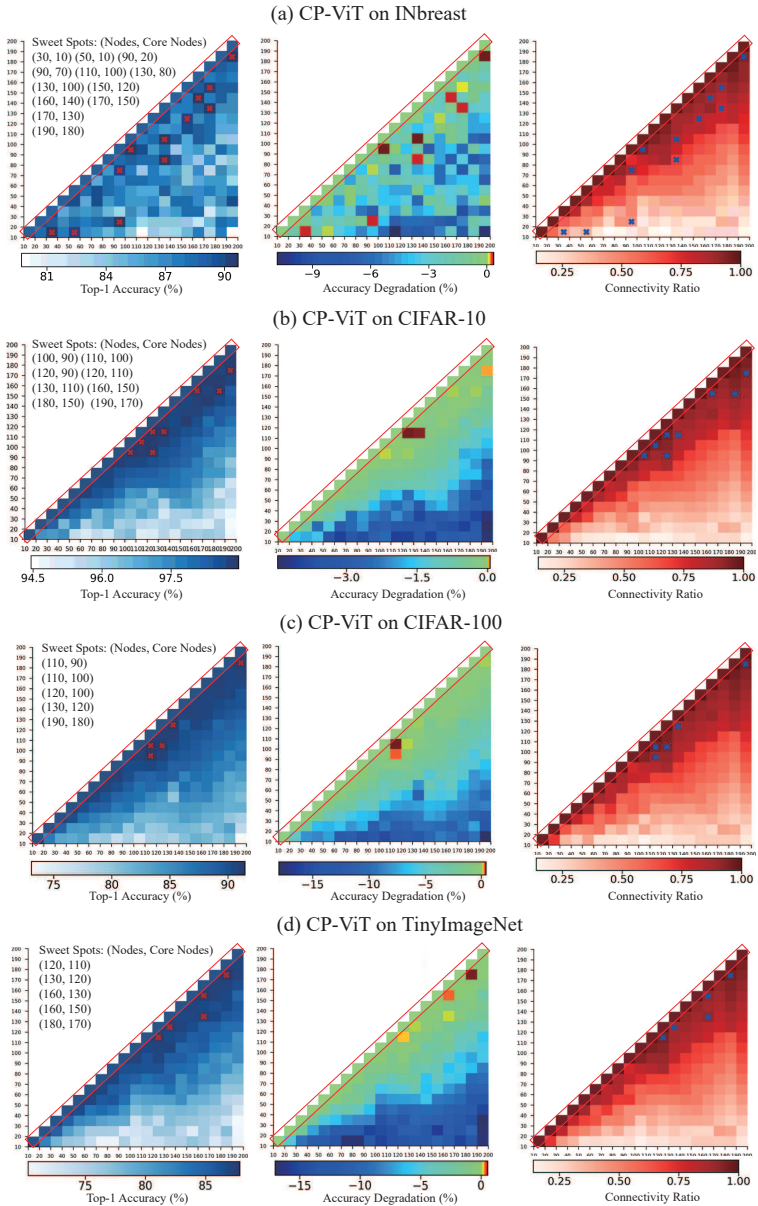}
\end{center}
\caption{Performance of CP-ViT measured using INbreast, CIFAR-10, CIFAR-100 and TinyImageNet datasets. Sub-figures on the left column under each datasets show the top 1 classification accuracy of the CP-ViTs and vanilla ViTs in the search space. A deeper color means higher top 1 accuracy. Sweet spots are marked by red crosses, in which CP-ViTs achieve better performance than vanilla ViT. Sub-figures on the middle column are the accuracy degradation of the CP-ViTs compared to vanilla ViTs. Sub-figures on the right column are the self-attention connection ratio of the CP-ViTs and vanilla ViT. Lighter color means a lower connection ratio. Sweet spots are marked by the blue crosses.}
\label{Performance}
\end{figure}

\begin{table}[!h]
\caption{Comparison between the proposed CP-ViT in sweet spots with finetuned vanilla ViT-S \cite{dosovitskiy2020image}. * means vanilla ViT-S finetuned by ourselves.}
\centering
\begin{tabular}{cccccc}
\hline
\multirow{1}{*} Dataset & Model  
                   &  CP Graph  & CR ($\%$) & $R_{cc}$,$R_{cp}$,$R_{pp}$ & Top1 Acc.($\%$)   \\
\hline
 & ViT-S(*)  &$(N,N)$    & $100.00$    &  $0.33,0.33,0.33$    &$89.91$       \\
 & CP-ViT   &$(30, 10)$    & $32.36$    &$0.58, 0.33, 0.09$   &  $90.58$\\
 & CP-ViT  &$(50,10)$   &  $\mathbf{29.20}$    & $0.53, 0.34, 0.12$    &  $90.01$\\
 & CP-ViT  &$(90, 20)$   & $43.82$    & $0.52, 0.36, 0.12$ &   $90.58$\\
 & CP-ViT   &$(90, 70)$    & $84.50$    &$0.54, 0.40, 0.06$   &  $90.01$\\
 & CP-ViT   &$(100, 90)$    & $92.80$    &$0.49, 0.39, 0.11$   &  $\mathbf{90.69}$\\
  INbreast & CP-ViT   &$(130, 80)$    & $31.34$    &$0.58, 0.34, 0.07$   &  $90.58$\\
 & CP-ViT   &$(130, 100)$    & $82.94$    &$0.57, 0.36, 0.07$   &  $\mathbf{90.69}$\\
 & CP-ViT   &$(150, 120)$    & $84.18$    &$0.57, 0.41, 0.02$   &  $90.01$\\
 & CP-ViT   &$(160, 140)$    & $87.77$    &$0.55, 0.41, 0.03$   &  $90.58$\\
 & CP-ViT   &$(170, 130)$    & $80.79$    &$0.57, 0.41, 0.02$   &  $90.58$\\
 & CP-ViT   &$(170, 150)$    & $87.65$    &$0.56, 0.41, 0.03$   &  $90.12$\\
 & CP-ViT   &$(190, 180)$    & $84.89$    &$0.52, 0.42, 0.05$   &  $90.69$\\
\hline
  & ViT-S(*) & $(N,N)$     &  $100.00$     & $0.33,0.33,0.33$ &   $98.50$  \\
 & CP-ViT &$(100, 90)$    & $92.80$    &$0.49, 0.39, 0.11$   &  $98.91$\\ 
 & CP-ViT  &$(110, 100)$    & $94.49$    &$0.53, 0.42, 0.05$   &  $98.91$\\
  & CP-ViT &$(120, 90) $    & $89.73$    &$0.51, 0.41, 0.08$   &  $98.91$\\
 CIFAR-10  & CP-ViT &$(120, 110)$    & $94.70$    &$0.49, 0.38, 0.12$   &  $98.97$\\
 & CP-ViT &$(130, 110)$    & $\mathbf{87.32}$    &$0.56, 0.40, 0.03$   &  $\mathbf{98.97}$\\
  & CP-ViT &$(160, 150) $    & $90.47$    &$0.54, 0.39, 0.06$   &  $98.91$\\
  & CP-ViT &$(180, 150) $    & $91.79$    &$0.50, 0.42, 0.07$   &  $98.91$\\
  & CP-ViT &$(190, 170) $    & $92.59$    &$0.53, 0.43, 0.03$   &  $98.94$\\
  \hline
 & ViT-S(*)  & $(N,N)$     &  $100.00$     & $0.33,0.33,0.33$ &   $91.10$  \\
 & CP-ViT &$(110, 90)$    & $88.96$    &$0.59, 0.37, 0.04$   &  $91.32$\\
 CIFAR-100 & CP-ViT  &$(110, 100)$    & $94.49$    &$0.53, 0.42, 0.05$   &  $\mathbf{91.45}$\\
  & CP-ViT &$(120, 100)$    & $92.40$    &$0.50, 0.41, 0.09$   &  $91.15$\\
  & CP-ViT &$(130, 120)$    & $\mathbf{87.50}$    &$0.58, 0.32, 0.09$   &  $91.11$\\
  & CP-ViT &$(190, 180)$    & $94.89$    &$0.52, 0.42, 0.05$   &  $91.12$\\
\hline
  & ViT-S(*)  &$(N,N)$    &  $100.00$     & $0.33,0.33,0.33$ &   $87.36$  \\
 & CP-ViT &$(120, 110)$    & $94.71$    &$0.49,0.39,0.12$   &  $87.51$\\
TinyImageNet & CP-ViT  &$(130, 120)$    & $\mathbf{87.50}$    &$0.58,0.33,0.09$   &  $87.37$\\
  & CP-ViT &$(160, 130)$    & $90.02$    &$0.54,0.44,0.02$   &  $87.40$\\
  & CP-ViT &$(160, 150)$    & $90.47$    &$0.54,0.40,0.06$   &  $87.63$\\
  & CP-ViT &$(180, 170)$    & $95.84$    &$0.50,0.43,0.07$   &  $\mathbf{87.84}$\\
\hline

\label{performance comparison}
\end{tabular}
\end{table}

We explored the performance of different types of CP graphs in the search space (Fig. \ref{CPGraphs}(a)) in terms of top $1$ accuracy and connection ratio. The connection ratio (CR) quantitatively measures the computational costs of different self-attention operations, which is defined by (\ref{6}):
\begin{equation}\label{6}
CR = \frac{  1_{M_{cp}} }{ \lVert{M_{cp}} \rVert_{1} }
\end{equation}
where $  1_{M_{cp}} $ represents the number of $1$s in the mask matrix of cp graphs - $  {M_{cp}} $ which is derived from the adjacency matrix of the CP graph, and ${\lVert \bullet \rVert}_{1}$ is the number of elements in the mask matrix. In general, CR represents the ratio of actual self-attention operations to the potential maximum self-attention operations. Given a graph, the potential maximum self-attention operation is fixed. Less actual self-attention operation means less computational cost and hence it has a smaller CR value. 

For each specific combination of different numbers of nodes/core nodes in the search space, we trained the CP-ViT with 5 different CP graph samples and reported the average result in Fig. \ref{Performance}. The four results in Fig. \ref{Performance}(a-d) correspond to four different datasets. For the results on each dataset, we display three subfigures: the top 1 accuracy (left), the accuracy degradation (middle), and the connection ratio (right). We highlighted the sweet spots, which are corresponding to the CP graphs that lead to improved performance\cite{you2020graph}, with red crosses in Fig. \ref{Performance}. In the top-1 accuracy of Fig. \ref{Performance}, deeper color means better performance. The accuracy degradation subfigures show the accuracy variation compared to fully connected self-attention ViTs. Our CP-ViTs gain a positive boost in sweep spots as it has higher accuracy than vanilla ViTs. At the same time, our CP-ViTs maintain competitive top-1 accuracy in most search space areas, as shown in the middle subfigures. The performance of CP-ViTs varies in the search space. This result indicates that different self-attention (wiring) patterns may have great influences on the performances of ViTs. Compared to vanilla ViTs with a fully-connected self-attention pattern, the proposed CP-ViT provides the potential for the model to only search for optimal self-attention patterns. The CRs of all the ViTs including vanilla ViTs and CP-ViTs were shown on the right. The CRs of the sweet spots were marked with a blue cross. Besides the improvement in classification accuracy ({$\verb+0.78+\%$} for INbreast, {$\verb+0.47+\%$} for CIFAR-10, {$\verb+0.35+\%$} for CIFAR-100, {$\verb+0.48+\%$} for TinyImageNet), the proposed CP-ViT also leads to a great reduction in connection ratio due to less self-attention operations ({$\verb+-70.80+\%$} connections for INbreast, {$\verb+-12.68+\%$} connections for CIFAR-10, {$\verb+-12.50+\%$} connections for CIFAR-100, {$\verb+-12.50+\%$} connections for TinyImageNet). The model setting, top $1$ accuracy, and CRs of different ViTs were reported in Table \ref{performance comparison}. For all the four datasets, our CP-ViT not only shows improved classification performance but also reduces connection ratio compared to vanilla ViTs. Interestingly, our results demonstrate that the “sweet spots" are corresponding to the wiring patterns (graphs) with CP structures, instead of fully connected self-attention.

We also compared the proposed CP-ViT with the state-of-the-art methods in Table \ref{Methods comparison}, including various convolutional networks and transformer architectures. Note that we applied the core-periphery principle to guide the design on small ViT, therefore, the counterparts we compared to in this work are also small-scale transformers and their variants.  “$--$" means there is no available reports or not applicable. As presented in the table, our method outperforms the CNNs, and a series of variants of transformers on these datasets, suggesting the superiority of the proposed CP-ViTs over the existing methods.

\begin{table}
\caption{Comparisons with state-of-the-art transformers and other architectures. }
\centering
\begin{tabular}{ccccc}
\hline
\multirow{1}{*} Model & CIFAR-10 & CIFAR-100 & TinyImageNet & INbreast   \\
\hline
ResNet-18\cite{he2016deep}   &$95.55$  &  $76.64$    &$67.33$ & 84.34   \\
ResNet-18+Gaze\cite{wang2022follow}   &$--$  &  $--$    &$--$  & 86.74 \\
ViT-S-SAM\cite{chen2021vision}   &$98.20$  &  $87.60$    &$87.50$   &  $90.20$ \\
ViT-S\cite{chen2021vision} &$97.60$   &   $85.70$  & $87.40$ & $89.91$   \\ 
DeiT-S\cite{touvron2021training}   &$97.50$   & $90.30$    &$86.90$   & $89.90$ \\
Mixer-S-SAM\cite{chen2021vision}   &$96.10$   & $82.40$    &$85.60$   & $87.60$ \\
T2T-ViT-12\cite{wang2021not}   &$98.53$   & $89.63$    &$86.20$   & $88.40$ \\
AutoFormer-S\cite{chen2021autoformer}   &$98.50$   & $90.60$& $87.60$   & $90.10$ \\
CP-ViT-S(ours)   &\bm{$98.97$}   &\bm{ $91.45$}  & \bm{ $87.84$}   & \bm{$90.69$}    \\
\hline
\label{Methods comparison}
\end{tabular}
\end{table}

\subsection{Visualization of Important Patches}
\begin{figure}[H]
\begin{center}
\includegraphics[scale=0.10]{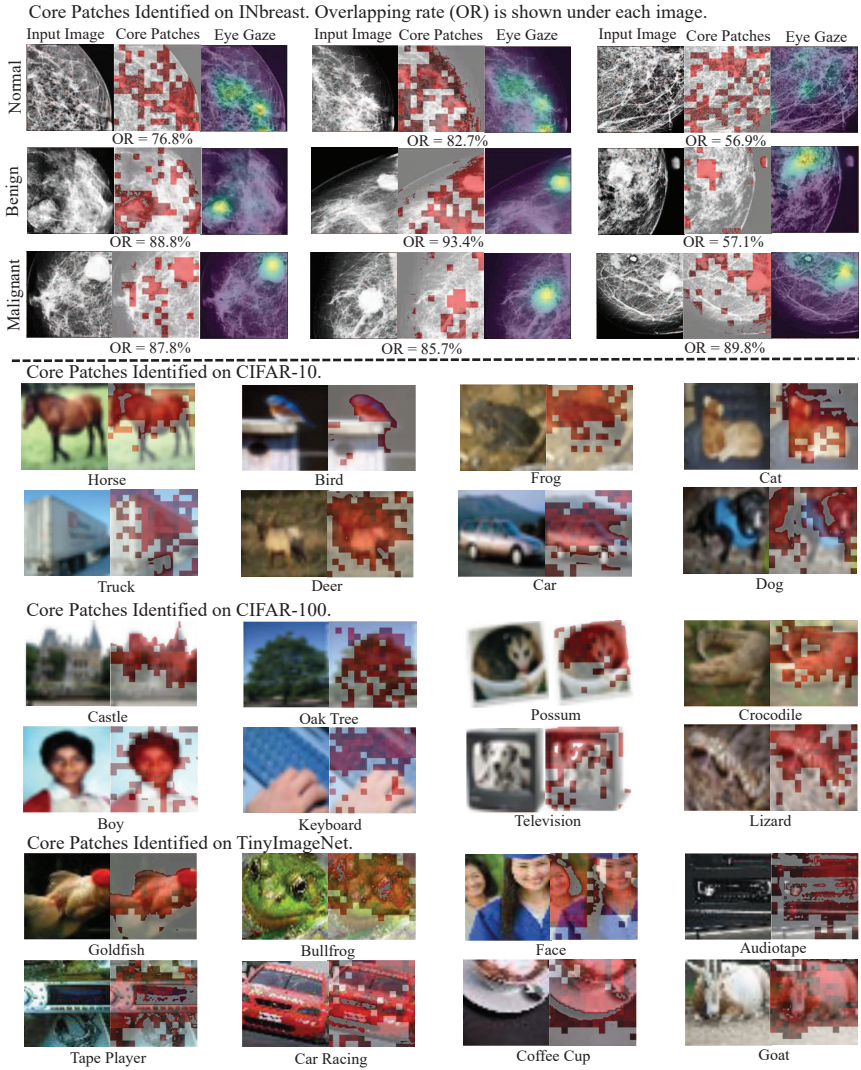}
\end{center}
\caption{Visualization of important image patches that were distributed to the core nodes. For the INbreast dataset (the first block), images of three randomly selected subjects for each class were shown. For each subject, there are three images displayed in three columns. The left column is the original image, the middle column shows the important patches marked by red, and the right column is the eye gaze of medical physicians on the image. For the natural image datasets (the second block, CIFAR-10, CIFAR-100 and TinyImageNet), the important patches identified in eight randomly selected classes were displayed. The left column is the original image, and the right column shows the identified core patches marked in red.}
\label{Importantpatches}
\end{figure}

Another advantage of CP-ViT is that it can potentially improve the interpretability of the deep-learning models via semi-intervention when linking the explainable concepts contained in the data to the instilled CP structures (section 3.2.3). In our CP-ViT the core nodes are expected to be associated with the important image patches relating to the classification tasks. To evaluate this, we show the patches that were redistributed to the core nodes when the model was well-trained in Fig.~\ref{Importantpatches}. For INBreast, we randomly selected the images of three subjects in each class and displayed the original images, the images overlaid with important patches, and the images overlaid with the expert's eye gazes in three columns. As shown in the Fig. \ref{Importantpatches}, the patches of the core nodes are well co-localized with the locations that were identified as diagnostic biomarkers of the disease in literature publications\cite{ibrokhimov2022two}. We also show the medical physicians' eye gaze maps on these images, given that the eye gaze acquired by eye-tracking equipment is considered the ground truth for identifying important areas in the image. The important patches identified by our CP-ViT highly overlap with the eye gaze maps, demonstrating the correspondence between the core nodes and the task-related concepts, i.e., the

\noindent important image patches. For natural image datasets, we also visualized the patches assigned to the core nodes under the black dotted line in Fig. \ref{Importantpatches}. It is clear that the objects in the patches of core nodes are semantically related to the class labels.

\begin{figure}
\begin{center}
\includegraphics[scale=0.12]{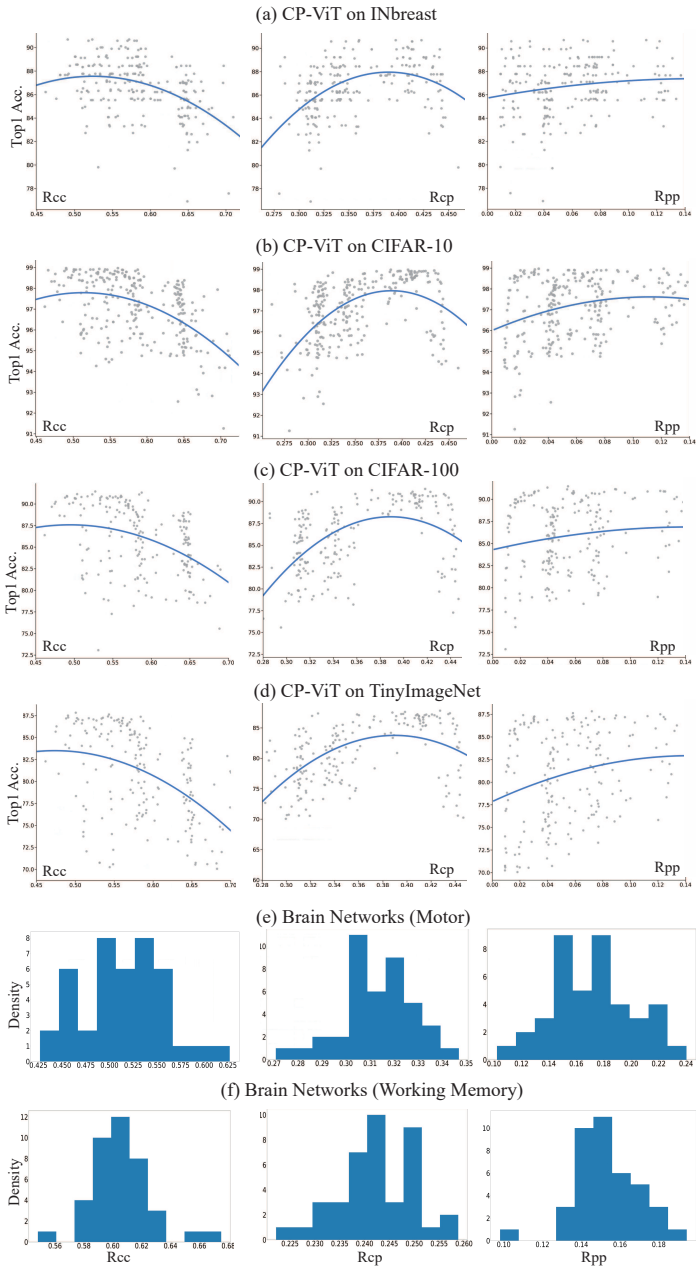}
\end{center}
\caption{Visualization of Core-periphery measures versus the classification performance. The regression results of the normalized independent probability versus the classification accuracy for experiments on each dataset are presented in (a), (b), (c), and (d). The core-periphery measures for brain networks of motor and working memory are shown in (e) and (f).}
\label{CPmeasures}
\end{figure}

\subsection{Fast Search for Sweet Spots}
Our proposed CP-ViT aims to achieve better performance more efficiently, by directly updating the initial dense wiring patterns with sparse CP graphs which are widely existing in BNN. Previous studies suggest that in ANN there exist  sweet spots that correspond to some specific wiring patterns leading to significantly improved performance\cite{you2020graph}. Therefore, it is interesting to investigate the relationship between sweet spots (the ANN structures with better performance) and the introduced CP structure. We conducted intensive experiments to illustrate how the accuracy changes under the CP measurements (in terms of normalized independent probability) and the results are summarized in Fig.~\ref{CPmeasures}. We found the normalized independent probabilities between core nodes - $R_{cc}$, core and periphery nodes - $R_{cp}$ and periphery nodes - $R_{pp}$ fall in different range: $[0.45, 0.70]$ for $R_{cc}$, $[0.25, 0.45]$ for $R_{cp}$, and $[0.00, 0.15]$ for $R_{cc}$. Both $R_{cc}$ and $R_{cp}$ display obvious and consistent patterns in terms of the relationship between ANN performance (accuracy) and CP properties: there exists a certain range of CP structures with which the corresponding wiring patterns of ANN can achieve better performance. For example, when the normalized independent probabilities between core and periphery nodes ($R_{cp}$) fall within the range of $[0.36, 0.42]$, our CP-ViT inclines to have the best accuracy on all four datasets. On the contrary, the normalized independent probabilities between periphery nodes ($R_{pp}$) show relatively less influence on the overall performance. These results suggest that the wiring patterns between core nodes and periphery nodes have more influence on the overall ANN performance than the wiring patterns between periphery nodes. For comparison, we also calculated the range of group-wise normalized independent probabilities in human functional brain networks when performing two different tasks - motor and working memory tasks. The results are shown in Fig.~\ref{CPmeasures} (e-f). Interestingly, the distribution of $R_{cc}$, $R_{cp}$ and $R_{cp}$ shows obvious overlaps among different functional brain networks though the major range of CP metrics is different from ANN (our CP-ViT). In general, our CP-ViT can leverage the CP structure to learn the optimal combinations of total nodes and core nodes, and to quickly find the sweet spots in a more efficient way.

\section{Methods}\label{sec11}

\subsection{Related Work}
\label{gen_inst}

\textbf{Core-periphery Structure} The Core-Periphery structure is a fundamental network signature that is composed of two qualitatively distinct components: a dense “core” of nodes strongly interconnected with one another, allowing for integrative information processing to facilitate the rapid transmission of the message, and a sparse “periphery” of nodes sparsely connected to the core and among each other\cite{gallagher2021clarified}. The Core-Periphery pattern has helped explain a broad range of phenomena in network-related domains, including online amplification\cite{barbera2015critical}, cognitive learning processes\cite{bassett2013task}, technological infrastructure organization\cite{alvarez2005k, carmi2007model}, and critical disease-spreading conduits\cite{kitsak2010identification}. All these phenomena suggest that the Core-Periphery pattern may play a critical role to ensure the effectiveness and efficiency of information exchange within the network. In the literature, there are two widely-used approaches for generating graphs with Core-Periphery property (CP graphs): the classic two-block model of Borgatti and Everett (BE algorithm)\cite{borgatti2000models} and the k-cores decomposition\cite{gallagher2021clarified}. The former approach partitions a network into a binary hub-and-spoke layout, while the latter one divides it into a layered hierarchy. In this work, for simplicity, we adopted a two-block model to generate a CP graph which is used to guide the self-attention operations between patches (tokens) in ViT. In this way, the Core-Periphery property is infused into the ViT model.

\textbf{Methods for Designing More Efficient ViT Architecture} ViT and its variants have achieved promising performances in various computer vision tasks, but their gigantic parameter counts, heavy run-time memory usage, and high computational cost become a major burden for the applications. Therefore, there is an urgent need to develop lightweight vision transformers with comparable performance and efficiency. For this purpose, several studies aimed to use network pruning, sparse training, and supernet-based NAS to slim vanilla ViT. \textbf{From token level}, Tang et al.\cite{tang2022patch} designed a patch slimming method to discard useless tokens. Evo-ViT\cite{xu2022evo} updated the selected informative and uninformative tokens with different computation paths. VTP\cite{zhu2021vision} reduced embedding dimensionality by introducing control coefficients. \textbf{From model architecture level}, UP-ViTs\cite{yu2021unified} pruned the channels in ViTs in a unified manner, including residual connections in all the blocks, multi-head self-attention (MHSA)\cite{vaswani2017attention}, feedforward neural layers (FFNs), normalization layers, and convolution layers in ViT variants. SViTE\cite{chen2021chasing} dynamically extracted and trained sparse subnetworks instead of training the entire model. To further co-explore data and architecture sparsity, a learnable token selector was used to determine the most vital image patch embeddings in the current input sample. AutoFormer\cite{chen2021autoformer} and ViTAS\cite{su2021vitas} leveraged supernet-based NAS to optimize the ViT architecture. Despite the remarkable improvements achieved by the above methods, both token-sampling and data-driven strategies may highly depend on the data and tasks performed, impeding the vision transformers’ generalization capability. A more universal principle (e.g., derived from BNNs) that can guide a more efficient design of ANN's architecture is much desired. In this work, we will leverage a widely existing Core-Periphery property in BNN to develop a more efficient CP-ViT.

\subsection{Core-Periphery Principle Guided Transformer}
\begin{figure}[H]
\begin{center}
\includegraphics[scale=0.35]{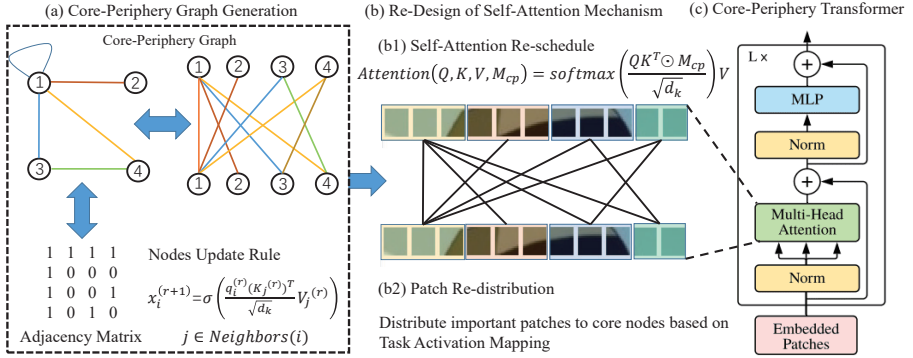}
\end{center}
\caption{Core-Periphery Principle Guided Re-design of Self-Attention. The proposed Core-Periphery guided re-design framework for ViTs consists of two major components: the Core-Periphery graph generator and the re-design of the self-attention mechanism. The basic idea is that we mapped the ViT structure to graphs and proposed a new graph representation paradigm to represent the self-attention mechanism. Under this paradigm, the design of the self-attention mechanism can be turned into a task of designing desirable graphs. (a) The CP graph generator was proposed to generate graphs with Core-Periphery property in a wide range of search spaces. (b) The self-attention of the nodes is controlled by the generated CP graph and the patches are re-distributed to different nodes by a novel patch distribution method. (c) The new self-attention mechanism will upgrade the regular self-attention in vanilla ViT. The new ViT architecture is thus named as CP-ViT.}
\label{model}
\end{figure}

The Core-Periphery principle can be applied to ViT and its variants via a unified framework that is illustrated in Fig. \ref{model}. The framework includes two main parts: Core-Periphery graph generation and Core-Periphery graph guided re-design of the self-attention mechanism.

\subsubsection{Core-Periphery Graph Generation}
The self-attention of our proposed CP-ViT is controlled by Core-Periphery graphs (CP graphs). We proposed a CP graph generator to generate a wide spectrum of CP graphs in the graph space defined by the number of total nodes and the core nodes. Although several graph generators have been proposed in previous works, they were not designed for generating CP graphs. For example, Erdos-Renyi (ER) generator samples graphs with given node and edge numbers uniformly and randomly\cite{erdos1960evolution}; Watts-Strogatz (WS) generator generates graphs with small-world properties\cite{watts1998collective}, and the complete graphs generator generates graphs where nodes are pair-wise densely connected with each other\cite{walker1992implementing}. 

To generate graphs with CP property, we proposed a novel CP graph generator that is parameterized by a total node number $n$, a core node number $m$, and three wiring thresholds $p_{cc}, \ p_{cp}, \ p_{pp}$ which are the wiring probabilities between the core-core nodes, core-periphery nodes, and periphery-periphery nodes, respectively. Based on these measures, the CP graph generation process is as follows: we first defined the core nodes number $m$ and the periphery nodes number $n-m$; Then, for each of the core-core node pairs, we used a random seed sampled from the continuous uniform distribution in $\left [ 0, 1 \right ] $ to generate a wiring probability $p_{rs}$. If the wiring probability is greater than the threshold $p_{cc}$, the two core nodes are connected. This wiring process is formulated as:
\begin{equation}\label{CPGenerationFormula}
A\left ( i ,j\right )=  \begin{cases}
1  & \text{ if } p_{rs}\ge p_{cc} \\
0  & \text{ if } p_{rs}<p_{cc}
\end{cases}
\end{equation}
where $A$ is the adjacency matrix of the generated graph, $1$ means that there exists an edge between the nodes $i$ and $j$, $0$ means there is no edge between the nodes. The same procedure was applied to core-periphery and periphery-periphery node pairs with the corresponding thresholds $\ p_{cp}$ and $\ p_{pp}$, respectively. In this way, by using different combinations of $n$, $m$, and wiring thresholds, we can generate a large number of candidate graphs in the graph space; finally, all the generated graphs were examined by the CP detection algorithm (BE algorithm)\cite{borgatti2000models} and the graphs with CP property will be used in the further steps to guide the self-attention operation.

\subsubsection{Core-Periphery Guided Self-Attention}
To instill the CP principle into the self-attention mechanism in ViT, we redesigned the self-attention operations according to the generated CP graphs: the patches are replaced by the nodes, and the new self-attention relations are replaced by the edges in the CP graph. Thus, the self-attention in the vanilla ViT can be represented as a complete graph, and similarly, the CP principle can be effectively and conveniently infused into the ViT architecture by upgrading the complete graph with the generated CP graphs. CP graph can be represented as $\mathcal{G} = ( \mathcal{V}, \mathcal{E})$, with nodes set $\mathcal{V}$ and edges set $\mathcal{E}$. The redesign of self-attention is formulated as:
\begin{equation}
x_{i}^{(r+1)}= \sigma^{(r)}( \{ (\frac{q_{i}^{(r)} (K_{j}^{(r)})^{T}  }{\sqrt{d_{k} } } )V_{j}^{(r)}, {\forall} j \in N(i) \} )
\end{equation}
where $\sigma(\cdot)$ is the activation function, which is usually the softmax function in ViTs, $q_{i} ^ {(r)} $ is the query of patches in the $i$-th node in $ \mathcal{G} $, $ N(i)= \{ i \| i \vee (i,j) \in \mathcal{E} \} $ are the neighborhood nodes of node $i$, $d_k$ is the dimension of queries and keys, and $K_{j}^{(r)}$ and $V_{j}^{(r)}$ are the key and value of patches in node $j$.

In vanilla ViT, one input image is divided into $ 196$ patches, and each patch resolution is 16 by 16. In CP-ViT, each node corresponds to a single patch or multiple patches. We proposed the following patch assignment pipeline to map the original patches to the nodes: for a CP graph with $n$ nodes, each node will be assigned to either $ \lfloor 196/n \rfloor +1 $ or $\lfloor 196/n \rfloor$ patches. For example, if we use a CP graph with $5$ nodes, the $5$ nodes will have 40, 39, 39, 39, and 39 patches, respectively; and if we use a CP graph with $196$ nodes, each node will correspond to a single patch. Note that the patches are randomly assigned to the nodes at the beginning of the training process, and then they will be re-distributed iteratively after each training epoch based on a novel patch distribution method that will be elaborated in the next section. Based on the above discussion, the CP graph-guided self-attention conducted at the node level can be formulated as:
\begin{equation}\label{2}
Attention(Q,K,V,M_{cp})=softmax(\frac{QK^T\odot M_{cp}}{\sqrt{d_k} }V )
\end{equation}
where the queries, keys, and values of all the patches are packed into the matrices $Q$, $K$, and $V$, respectively. $M_{cp}$ is the mask matrix derived from the adjacency matrix $A$ of the CP graph, and $\odot$ is the dot product. The size of the mask matrix $M_{cp}$ is $197 \times 197$ (196 patches plus 1 classification token), and it is a symmetric matrix. The derivation process of $M_{cp}$ is as follows: for a CP graph with 5 nodes, the 5 nodes have 40, 39, 39, 39, and 39 patches, respectively. If the element $(1,2)$ in the corresponding adjacency $A$ is 1, which means the node \#$1$ is connecting to the node \#$2$, and as a result, the $40$ patches corresponding to the node \#$1$ are connecting to the $39$ patches associated with the node \#$2$. Therefore, the elements at $(1:40, 40:79)$ and $(40:79, 1:40)$ in the mask matrix $M_{cp}$ will be $1$, where the $(40:79, 1:40)$ means the elements from the $40$th row to $79$th row, and from the $1$st column to the $40$th column. The elements in the last row and column of $M_{cp}$ are 1 because the classification token is connected to all the nodes, including both core and periphery nodes. Similar to the multi-head attention in transformers\cite{vaswani2017attention}, our proposed CP multi-head attention is formulated as:
\begin{equation}
\begin{aligned}
MultiHead(Q,K,V,M_{cp})=Concat(head_{1}, ...,head_{h})W^{o}\\
where \  head_{i}=Attention(  QW_{i}^{Q}, KW_{i}^{K} , VW_{i}^{V},M_{cp} ) 
\end{aligned}
\end{equation}
where the parameter matrices $W_{i}^{Q}$, $W_{i}^{K}$, $W_{i}^{V}$ and $W^{O}$ are the projections. Multi-head attention helps the model to jointly aggregate information from different representation subspaces at various positions. In this work, we apply the CP principle to each representation subspace.

\renewcommand{\algorithmicrequire}{ \textbf{Input:}}     
\renewcommand{\algorithmicensure}{ \textbf{Output:}}    
\begin{algorithm}
\caption{Patch Re-Distribution}\label{PatchRedisAlgo}
\begin{algorithmic}[1]
\Require Likelihoods of an image belonging to a particular class (before activation layer) $y$, patch embeddings $P^k, k = 1,2,...,196$.
\State Calculate the gradients of the likelihoods $y$
with respect to patch embeddings $P^{k}$, respectively.  $\frac{\partial y}{\partial P^{k}_{i}}$.
\State Obtain patch important weights $\alpha _{k}, k = 1,2,...,196$ by average-pooling of gradients over the feature dimension, $\alpha _{k} = \frac{1}{Z} \sum_{i=1}^{Z} \frac{\partial y}{\partial P^{k}_{i}}$, where $Z$ is the dimension of the patch embeddings. 
\State Sort the patch important weights $\alpha _{k}$ in a descending manner, $Sort(\alpha _{k})$.
\State Determine the number of patches assigned to core nodes, for simplicity, we call these patches as core patches.
\State Match the core patches to core nodes in a way that the patches with higher importance weights are distributed to the core nodes with a higher degree.
\State Re-organized the patches of the images according to the importance weights.
\Ensure Patch re-organized images.
\end{algorithmic}
\end{algorithm}
\bigskip

\begin{figure}[H]
\begin{center}
\includegraphics[scale=0.43]{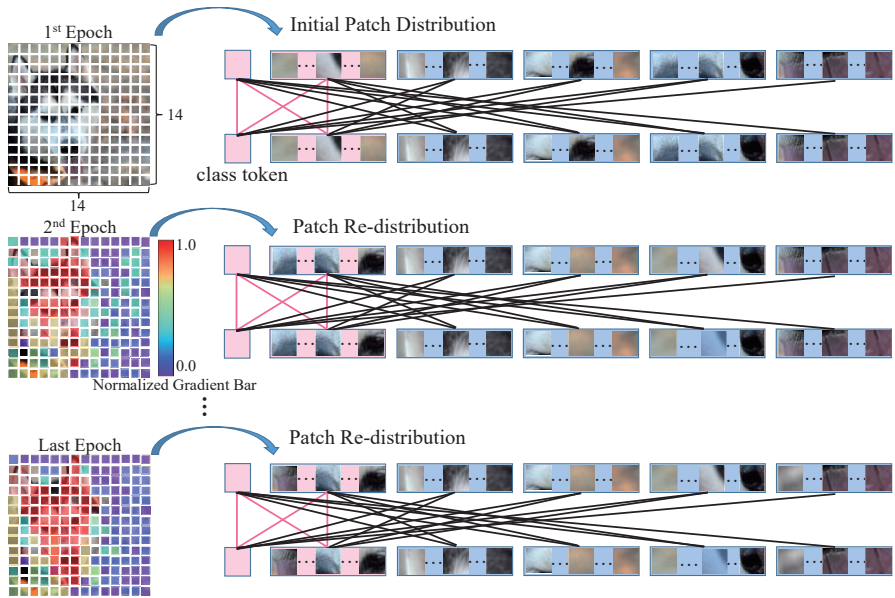}
\end{center}
\caption{Illustration of Patch Redistribution Process. The pink nodes are the core nodes, while the blue nodes are the periphery nodes. The initial patch distribution at the first epoch is the same as the vanilla ViTs. After each iteration during the training process, the gradients of patches discriminate from each other due to different contributions to the classification. The red the image patches are, the high gradient they are. Thus, the core patches that contribute most to the classification task are re-distributed to core nodes. }
\label{patchRedistribution}
\end{figure}

\subsubsection{Patch Redistribution}
The CP structure inclines to make the communication and message exchange at core nodes more intensive while less frequent among periphery nodes. This is based on the fact that the core nodes usually process the most important information in many biological networks\cite{bassett2013task}. To this end, we need to evaluate the importance of the patches and select the most important ones to assign to the core nodes, which is defined as task-related activation feature mapping. For a specific task of CP-ViT, in order to identify the important patches, we computed the gradients of the output $y$ (before the activation function) with respect to patch features (after patch embedding) $P^k$, i.e. $\frac{\partial y }{\partial P^{k}} $. These gradients flowing back to the patch features are global-average-pooling over the feature dimensions to obtain the patch importance weights. The important weights are:
\begin{equation}\label{4}
\alpha _{k} = \frac{1}{Z} \sum_{i=1}^{Z} \frac{\partial y}{\partial P^{k}_{i}}
\end{equation}
where $Z$ is the dimension of the patch embedding features. After we have the weights of all the patches, the top $K$ patches that have the highest weights are selected and re-distributed to the core nodes. Note that the patch distribution process is not random but distributed based on the nodes' degree in a in a descending manner: the patches with higher importance weights are distributed to the core nodes with a higher degree. The algorithm for patch redistribution is detailed described in algorithm \ref{PatchRedisAlgo}, and the corresponding patch redistribution process is illustrated in Fig. \ref{patchRedistribution}. As shown in Fig. \ref{patchRedistribution}, the image patches were randomly distributed at the first epoch but as the training process proceeded, patches with high gradients are identified as important patches and gradually redistributed to the core nodes. After certain iteration epochs, those patches that contribute the most to the classification result will be distributed to the core nodes.

\section{Conclusion}\label{sec5}

In this work, we proactively instilled an organizational principle of BNN, that is, Core-Periphery property, to guide the design of ANN of ViT. For this, we provide a unified framework to introduce the core-periphery principle to guide the design of self-attention, the most prominent mechanism in transformers. Our extensive experiments suggest that there exist sweet spots of CP graphs that lead to CP-ViTs with significantly improved predictive performance. In general, our work advances the state of the art in three ways: 1) this work provides novel insights for brain-inspired AI by applying organizational principles of BNNs to ANN design; 2) the optimized CP-ViT can significantly improve its predictive performance while have the potential to reduce the unnecessary computational cost; and 3) the core nodes in CP-ViT are associated with task-related meaningful image patches, which can significantly enhance the interpretability of the trained deep model.

\bmhead{Acknowledgments}

This work was supported by the National Institute On Aging of the National Institutes of Health under Award Number R01AG075582 and the National Institute Of Neurological Disorders And Stroke of the National Institutes of Health under Award Number RF1NS128534.

\bibliography{sn-bibliography}


\begin{thebibliography}{47}
\ifx \bisbn   \undefined \def \bisbn  #1{ISBN #1}\fi
\ifx \binits  \undefined \def \binits#1{#1}\fi
\ifx \bauthor  \undefined \def \bauthor#1{#1}\fi
\ifx \batitle  \undefined \def \batitle#1{#1}\fi
\ifx \bjtitle  \undefined \def \bjtitle#1{#1}\fi
\ifx \bvolume  \undefined \def \bvolume#1{\textbf{#1}}\fi
\ifx \byear  \undefined \def \byear#1{#1}\fi
\ifx \bissue  \undefined \def \bissue#1{#1}\fi
\ifx \bfpage  \undefined \def \bfpage#1{#1}\fi
\ifx \blpage  \undefined \def \blpage #1{#1}\fi
\ifx \burl  \undefined \def \burl#1{\textsf{#1}}\fi
\ifx \doiurl  \undefined \def \doiurl#1{\url{https://doi.org/#1}}\fi
\ifx \betal  \undefined \def \betal{\textit{et al.}}\fi
\ifx \binstitute  \undefined \def \binstitute#1{#1}\fi
\ifx \binstitutionaled  \undefined \def \binstitutionaled#1{#1}\fi
\ifx \bctitle  \undefined \def \bctitle#1{#1}\fi
\ifx \beditor  \undefined \def \beditor#1{#1}\fi
\ifx \bpublisher  \undefined \def \bpublisher#1{#1}\fi
\ifx \bbtitle  \undefined \def \bbtitle#1{#1}\fi
\ifx \bedition  \undefined \def \bedition#1{#1}\fi
\ifx \bseriesno  \undefined \def \bseriesno#1{#1}\fi
\ifx \blocation  \undefined \def \blocation#1{#1}\fi
\ifx \bsertitle  \undefined \def \bsertitle#1{#1}\fi
\ifx \bsnm \undefined \def \bsnm#1{#1}\fi
\ifx \bsuffix \undefined \def \bsuffix#1{#1}\fi
\ifx \bparticle \undefined \def \bparticle#1{#1}\fi
\ifx \barticle \undefined \def \barticle#1{#1}\fi
\bibcommenthead
\ifx \bconfdate \undefined \def \bconfdate #1{#1}\fi
\ifx \botherref \undefined \def \botherref #1{#1}\fi
\ifx \url \undefined \def \url#1{\textsf{#1}}\fi
\ifx \bchapter \undefined \def \bchapter#1{#1}\fi
\ifx \bbook \undefined \def \bbook#1{#1}\fi
\ifx \bcomment \undefined \def \bcomment#1{#1}\fi
\ifx \oauthor \undefined \def \oauthor#1{#1}\fi
\ifx \citeauthoryear \undefined \def \citeauthoryear#1{#1}\fi
\ifx \endbibitem  \undefined \def \endbibitem {}\fi
\ifx \bconflocation  \undefined \def \bconflocation#1{#1}\fi
\ifx \arxivurl  \undefined \def \arxivurl#1{\textsf{#1}}\fi
\csname PreBibitemsHook\endcsname

\bibitem{he2016deep}
\begin{botherref}
\oauthor{\bsnm{He}, \binits{K.}},
\oauthor{\bsnm{Zhang}, \binits{X.}},
\oauthor{\bsnm{Ren}, \binits{S.}},
\oauthor{\bsnm{Sun}, \binits{J.}}:
Deep residual learning for image recognition. cvpr. 2016
(2016)
\end{botherref}
\endbibitem

\bibitem{krizhevsky2017imagenet}
\begin{barticle}
\bauthor{\bsnm{Krizhevsky}, \binits{A.}},
\bauthor{\bsnm{Sutskever}, \binits{I.}},
\bauthor{\bsnm{Hinton}, \binits{G.E.}}:
\batitle{Imagenet classification with deep convolutional neural networks}.
\bjtitle{Communications of the ACM}
\bvolume{60}(\bissue{6}),
\bfpage{84}--\blpage{90}
(\byear{2017})
\end{barticle}
\endbibitem

\bibitem{vaswani2017attention}
\begin{botherref}
\oauthor{\bsnm{Vaswani}, \binits{A.}},
\oauthor{\bsnm{Shazeer}, \binits{N.}},
\oauthor{\bsnm{Parmar}, \binits{N.}},
\oauthor{\bsnm{Uszkoreit}, \binits{J.}},
\oauthor{\bsnm{Jones}, \binits{L.}},
\oauthor{\bsnm{Gomez}, \binits{A.N.}},
\oauthor{\bsnm{Kaiser}, \binits{{\L}.}},
\oauthor{\bsnm{Polosukhin}, \binits{I.}}:
Attention is all you need.
Advances in neural information processing systems
\textbf{30}
(2017)
\end{botherref}
\endbibitem

\bibitem{sherstinsky2020fundamentals}
\begin{barticle}
\bauthor{\bsnm{Sherstinsky}, \binits{A.}}:
\batitle{Fundamentals of recurrent neural network (rnn) and long short-term
  memory (lstm) network}.
\bjtitle{Physica D: Nonlinear Phenomena}
\bvolume{404},
\bfpage{132306}
(\byear{2020})
\end{barticle}
\endbibitem

\bibitem{tealab2018time}
\begin{barticle}
\bauthor{\bsnm{Tealab}, \binits{A.}}:
\batitle{Time series forecasting using artificial neural networks
  methodologies: A systematic review}.
\bjtitle{Future Computing and Informatics Journal}
\bvolume{3}(\bissue{2}),
\bfpage{334}--\blpage{340}
(\byear{2018})
\end{barticle}
\endbibitem

\bibitem{devlin2018bert}
\begin{botherref}
\oauthor{\bsnm{Devlin}, \binits{J.}},
\oauthor{\bsnm{Chang}, \binits{M.-W.}},
\oauthor{\bsnm{Lee}, \binits{K.}},
\oauthor{\bsnm{Toutanova}, \binits{K.}}:
Bert: Pre-training of deep bidirectional transformers for language
  understanding.
arXiv preprint arXiv:1810.04805
(2018)
\end{botherref}
\endbibitem

\bibitem{dosovitskiy2020image}
\begin{botherref}
\oauthor{\bsnm{Dosovitskiy}, \binits{A.}},
\oauthor{\bsnm{Beyer}, \binits{L.}},
\oauthor{\bsnm{Kolesnikov}, \binits{A.}},
\oauthor{\bsnm{Weissenborn}, \binits{D.}},
\oauthor{\bsnm{Zhai}, \binits{X.}},
\oauthor{\bsnm{Unterthiner}, \binits{T.}},
\oauthor{\bsnm{Dehghani}, \binits{M.}},
\oauthor{\bsnm{Minderer}, \binits{M.}},
\oauthor{\bsnm{Heigold}, \binits{G.}},
\oauthor{\bsnm{Gelly}, \binits{S.}}, et al.:
An image is worth 16x16 words: Transformers for image recognition at scale.
arXiv preprint arXiv:2010.11929
(2020)
\end{botherref}
\endbibitem

\bibitem{liu2021swin}
\begin{bchapter}
\bauthor{\bsnm{Liu}, \binits{Z.}},
\bauthor{\bsnm{Lin}, \binits{Y.}},
\bauthor{\bsnm{Cao}, \binits{Y.}},
\bauthor{\bsnm{Hu}, \binits{H.}},
\bauthor{\bsnm{Wei}, \binits{Y.}},
\bauthor{\bsnm{Zhang}, \binits{Z.}},
\bauthor{\bsnm{Lin}, \binits{S.}},
\bauthor{\bsnm{Guo}, \binits{B.}}:
\bctitle{Swin transformer: Hierarchical vision transformer using shifted
  windows}.
In: \bbtitle{Proceedings of the IEEE/CVF International Conference on Computer
  Vision},
pp. \bfpage{10012}--\blpage{10022}
(\byear{2021})
\end{bchapter}
\endbibitem

\bibitem{tolstikhin2021mlp}
\begin{barticle}
\bauthor{\bsnm{Tolstikhin}, \binits{I.O.}},
\bauthor{\bsnm{Houlsby}, \binits{N.}},
\bauthor{\bsnm{Kolesnikov}, \binits{A.}},
\bauthor{\bsnm{Beyer}, \binits{L.}},
\bauthor{\bsnm{Zhai}, \binits{X.}},
\bauthor{\bsnm{Unterthiner}, \binits{T.}},
\bauthor{\bsnm{Yung}, \binits{J.}},
\bauthor{\bsnm{Steiner}, \binits{A.}},
\bauthor{\bsnm{Keysers}, \binits{D.}},
\bauthor{\bsnm{Uszkoreit}, \binits{J.}}, \betal:
\batitle{Mlp-mixer: An all-mlp architecture for vision}.
\bjtitle{Advances in Neural Information Processing Systems}
\bvolume{34},
\bfpage{24261}--\blpage{24272}
(\byear{2021})
\end{barticle}
\endbibitem

\bibitem{yu2022metaformer}
\begin{bchapter}
\bauthor{\bsnm{Yu}, \binits{W.}},
\bauthor{\bsnm{Luo}, \binits{M.}},
\bauthor{\bsnm{Zhou}, \binits{P.}},
\bauthor{\bsnm{Si}, \binits{C.}},
\bauthor{\bsnm{Zhou}, \binits{Y.}},
\bauthor{\bsnm{Wang}, \binits{X.}},
\bauthor{\bsnm{Feng}, \binits{J.}},
\bauthor{\bsnm{Yan}, \binits{S.}}:
\bctitle{Metaformer is actually what you need for vision}.
In: \bbtitle{Proceedings of the IEEE/CVF Conference on Computer Vision and
  Pattern Recognition},
pp. \bfpage{10819}--\blpage{10829}
(\byear{2022})
\end{bchapter}
\endbibitem

\bibitem{zoph2016neural}
\begin{botherref}
\oauthor{\bsnm{Zoph}, \binits{B.}},
\oauthor{\bsnm{Le}, \binits{Q.V.}}:
Neural architecture search with reinforcement learning.
arXiv preprint arXiv:1611.01578
(2016)
\end{botherref}
\endbibitem

\bibitem{ren2021comprehensive}
\begin{barticle}
\bauthor{\bsnm{Ren}, \binits{P.}},
\bauthor{\bsnm{Xiao}, \binits{Y.}},
\bauthor{\bsnm{Chang}, \binits{X.}},
\bauthor{\bsnm{Huang}, \binits{P.-Y.}},
\bauthor{\bsnm{Li}, \binits{Z.}},
\bauthor{\bsnm{Chen}, \binits{X.}},
\bauthor{\bsnm{Wang}, \binits{X.}}:
\batitle{A comprehensive survey of neural architecture search: Challenges and
  solutions}.
\bjtitle{ACM Computing Surveys (CSUR)}
\bvolume{54}(\bissue{4}),
\bfpage{1}--\blpage{34}
(\byear{2021})
\end{barticle}
\endbibitem

\bibitem{elsken2019neural}
\begin{barticle}
\bauthor{\bsnm{Elsken}, \binits{T.}},
\bauthor{\bsnm{Metzen}, \binits{J.H.}},
\bauthor{\bsnm{Hutter}, \binits{F.}}:
\batitle{Neural architecture search: A survey}.
\bjtitle{The Journal of Machine Learning Research}
\bvolume{20}(\bissue{1}),
\bfpage{1997}--\blpage{2017}
(\byear{2019})
\end{barticle}
\endbibitem

\bibitem{zhang2021explainable}
\begin{barticle}
\bauthor{\bsnm{Zhang}, \binits{Y.}},
\bauthor{\bsnm{Choi}, \binits{M.}},
\bauthor{\bsnm{Han}, \binits{K.}},
\bauthor{\bsnm{Liu}, \binits{Z.}}:
\batitle{Explainable semantic space by grounding language to vision with
  cross-modal contrastive learning}.
\bjtitle{Advances in Neural Information Processing Systems}
\bvolume{34},
\bfpage{18513}--\blpage{18526}
(\byear{2021})
\end{barticle}
\endbibitem

\bibitem{you2020graph}
\begin{bchapter}
\bauthor{\bsnm{You}, \binits{J.}},
\bauthor{\bsnm{Leskovec}, \binits{J.}},
\bauthor{\bsnm{He}, \binits{K.}},
\bauthor{\bsnm{Xie}, \binits{S.}}:
\bctitle{Graph structure of neural networks}.
In: \bbtitle{International Conference on Machine Learning},
pp. \bfpage{10881}--\blpage{10891}
(\byear{2020}).
\bcomment{PMLR}
\end{bchapter}
\endbibitem

\bibitem{bassett2013task}
\begin{barticle}
\bauthor{\bsnm{Bassett}, \binits{D.S.}},
\bauthor{\bsnm{Wymbs}, \binits{N.F.}},
\bauthor{\bsnm{Rombach}, \binits{M.P.}},
\bauthor{\bsnm{Porter}, \binits{M.A.}},
\bauthor{\bsnm{Mucha}, \binits{P.J.}},
\bauthor{\bsnm{Grafton}, \binits{S.T.}}:
\batitle{Task-based core-periphery organization of human brain dynamics}.
\bjtitle{PLoS computational biology}
\bvolume{9}(\bissue{9}),
\bfpage{1003171}
(\byear{2013})
\end{barticle}
\endbibitem

\bibitem{gu2020unifying}
\begin{barticle}
\bauthor{\bsnm{Gu}, \binits{S.}},
\bauthor{\bsnm{Xia}, \binits{C.H.}},
\bauthor{\bsnm{Ciric}, \binits{R.}},
\bauthor{\bsnm{Moore}, \binits{T.M.}},
\bauthor{\bsnm{Gur}, \binits{R.C.}},
\bauthor{\bsnm{Gur}, \binits{R.E.}},
\bauthor{\bsnm{Satterthwaite}, \binits{T.D.}},
\bauthor{\bsnm{Bassett}, \binits{D.S.}}:
\batitle{Unifying the notions of modularity and core--periphery structure in
  functional brain networks during youth}.
\bjtitle{Cerebral Cortex}
\bvolume{30}(\bissue{3}),
\bfpage{1087}--\blpage{1102}
(\byear{2020})
\end{barticle}
\endbibitem

\bibitem{van2013wu}
\begin{barticle}
\bauthor{\bsnm{Van~Essen}, \binits{D.C.}},
\bauthor{\bsnm{Smith}, \binits{S.M.}},
\bauthor{\bsnm{Barch}, \binits{D.M.}},
\bauthor{\bsnm{Behrens}, \binits{T.E.}},
\bauthor{\bsnm{Yacoub}, \binits{E.}},
\bauthor{\bsnm{Ugurbil}, \binits{K.}},
\bauthor{\bsnm{Consortium}, \binits{W.-M.H.}}, \betal:
\batitle{The wu-minn human connectome project: an overview}.
\bjtitle{Neuroimage}
\bvolume{80},
\bfpage{62}--\blpage{79}
(\byear{2013})
\end{barticle}
\endbibitem

\bibitem{cucuringu2016detection}
\begin{barticle}
\bauthor{\bsnm{Cucuringu}, \binits{M.}},
\bauthor{\bsnm{Rombach}, \binits{P.}},
\bauthor{\bsnm{Lee}, \binits{S.H.}},
\bauthor{\bsnm{Porter}, \binits{M.A.}}:
\batitle{Detection of core--periphery structure in networks using spectral
  methods and geodesic paths}.
\bjtitle{European Journal of Applied Mathematics}
\bvolume{27}(\bissue{6}),
\bfpage{846}--\blpage{887}
(\byear{2016})
\end{barticle}
\endbibitem

\bibitem{holme2005core}
\begin{barticle}
\bauthor{\bsnm{Holme}, \binits{P.}}:
\batitle{Core-periphery organization of complex networks}.
\bjtitle{Physical Review E}
\bvolume{72}(\bissue{4}),
\bfpage{046111}
(\byear{2005})
\end{barticle}
\endbibitem

\bibitem{rombach2014core}
\begin{barticle}
\bauthor{\bsnm{Rombach}, \binits{M.P.}},
\bauthor{\bsnm{Porter}, \binits{M.A.}},
\bauthor{\bsnm{Fowler}, \binits{J.H.}},
\bauthor{\bsnm{Mucha}, \binits{P.J.}}:
\batitle{Core-periphery structure in networks}.
\bjtitle{SIAM Journal on Applied mathematics}
\bvolume{74}(\bissue{1}),
\bfpage{167}--\blpage{190}
(\byear{2014})
\end{barticle}
\endbibitem

\bibitem{liu2019cerebral}
\begin{barticle}
\bauthor{\bsnm{Liu}, \binits{H.}},
\bauthor{\bsnm{Zhang}, \binits{S.}},
\bauthor{\bsnm{Jiang}, \binits{X.}},
\bauthor{\bsnm{Zhang}, \binits{T.}},
\bauthor{\bsnm{Huang}, \binits{H.}},
\bauthor{\bsnm{Ge}, \binits{F.}},
\bauthor{\bsnm{Zhao}, \binits{L.}},
\bauthor{\bsnm{Li}, \binits{X.}},
\bauthor{\bsnm{Hu}, \binits{X.}},
\bauthor{\bsnm{Han}, \binits{J.}}, \betal:
\batitle{The cerebral cortex is bisectionally segregated into two fundamentally
  different functional units of gyri and sulci}.
\bjtitle{Cerebral Cortex}
\bvolume{29}(\bissue{10}),
\bfpage{4238}--\blpage{4252}
(\byear{2019})
\end{barticle}
\endbibitem

\bibitem{chen2021vision}
\begin{botherref}
\oauthor{\bsnm{Chen}, \binits{X.}},
\oauthor{\bsnm{Hsieh}, \binits{C.-J.}},
\oauthor{\bsnm{Gong}, \binits{B.}}:
When vision transformers outperform resnets without pre-training or strong data
  augmentations.
arXiv preprint arXiv:2106.01548
(2021)
\end{botherref}
\endbibitem

\bibitem{moreira2012inbreast}
\begin{barticle}
\bauthor{\bsnm{Moreira}, \binits{I.C.}},
\bauthor{\bsnm{Amaral}, \binits{I.}},
\bauthor{\bsnm{Domingues}, \binits{I.}},
\bauthor{\bsnm{Cardoso}, \binits{A.}},
\bauthor{\bsnm{Cardoso}, \binits{M.J.}},
\bauthor{\bsnm{Cardoso}, \binits{J.S.}}:
\batitle{Inbreast: toward a full-field digital mammographic database}.
\bjtitle{Academic radiology}
\bvolume{19}(\bissue{2}),
\bfpage{236}--\blpage{248}
(\byear{2012})
\end{barticle}
\endbibitem

\bibitem{krizhevsky2009learning}
\begin{botherref}
\oauthor{\bsnm{Krizhevsky}, \binits{A.}},
\oauthor{\bsnm{Hinton}, \binits{G.}}, et al.:
Learning multiple layers of features from tiny images
(2009)
\end{botherref}
\endbibitem

\bibitem{griffin2007caltech}
\begin{botherref}
\oauthor{\bsnm{Griffin}, \binits{G.}},
\oauthor{\bsnm{Holub}, \binits{A.}},
\oauthor{\bsnm{Perona}, \binits{P.}}:
Caltech-256 object category dataset
(2007)
\end{botherref}
\endbibitem

\bibitem{loshchilov2016sgdr}
\begin{botherref}
\oauthor{\bsnm{Loshchilov}, \binits{I.}},
\oauthor{\bsnm{Hutter}, \binits{F.}}:
Sgdr: Stochastic gradient descent with warm restarts.
arXiv preprint arXiv:1608.03983
(2016)
\end{botherref}
\endbibitem

\bibitem{wang2022follow}
\begin{botherref}
\oauthor{\bsnm{Wang}, \binits{S.}},
\oauthor{\bsnm{Ouyang}, \binits{X.}},
\oauthor{\bsnm{Liu}, \binits{T.}},
\oauthor{\bsnm{Wang}, \binits{Q.}},
\oauthor{\bsnm{Shen}, \binits{D.}}:
Follow my eye: Using gaze to supervise computer-aided diagnosis.
IEEE Transactions on Medical Imaging
(2022)
\end{botherref}
\endbibitem

\bibitem{touvron2021training}
\begin{bchapter}
\bauthor{\bsnm{Touvron}, \binits{H.}},
\bauthor{\bsnm{Cord}, \binits{M.}},
\bauthor{\bsnm{Douze}, \binits{M.}},
\bauthor{\bsnm{Massa}, \binits{F.}},
\bauthor{\bsnm{Sablayrolles}, \binits{A.}},
\bauthor{\bsnm{J{\'e}gou}, \binits{H.}}:
\bctitle{Training data-efficient image transformers \& distillation through
  attention}.
In: \bbtitle{International Conference on Machine Learning},
pp. \bfpage{10347}--\blpage{10357}
(\byear{2021}).
\bcomment{PMLR}
\end{bchapter}
\endbibitem

\bibitem{wang2021not}
\begin{barticle}
\bauthor{\bsnm{Wang}, \binits{Y.}},
\bauthor{\bsnm{Huang}, \binits{R.}},
\bauthor{\bsnm{Song}, \binits{S.}},
\bauthor{\bsnm{Huang}, \binits{Z.}},
\bauthor{\bsnm{Huang}, \binits{G.}}:
\batitle{Not all images are worth 16x16 words: Dynamic transformers for
  efficient image recognition}.
\bjtitle{Advances in Neural Information Processing Systems}
\bvolume{34},
\bfpage{11960}--\blpage{11973}
(\byear{2021})
\end{barticle}
\endbibitem

\bibitem{chen2021autoformer}
\begin{bchapter}
\bauthor{\bsnm{Chen}, \binits{M.}},
\bauthor{\bsnm{Peng}, \binits{H.}},
\bauthor{\bsnm{Fu}, \binits{J.}},
\bauthor{\bsnm{Ling}, \binits{H.}}:
\bctitle{Autoformer: Searching transformers for visual recognition}.
In: \bbtitle{Proceedings of the IEEE/CVF International Conference on Computer
  Vision},
pp. \bfpage{12270}--\blpage{12280}
(\byear{2021})
\end{bchapter}
\endbibitem

\bibitem{ibrokhimov2022two}
\begin{barticle}
\bauthor{\bsnm{Ibrokhimov}, \binits{B.}},
\bauthor{\bsnm{Kang}, \binits{J.-Y.}}:
\batitle{Two-stage deep learning method for breast cancer detection using
  high-resolution mammogram images}.
\bjtitle{Applied Sciences}
\bvolume{12}(\bissue{9}),
\bfpage{4616}
(\byear{2022})
\end{barticle}
\endbibitem

\bibitem{gallagher2021clarified}
\begin{barticle}
\bauthor{\bsnm{Gallagher}, \binits{R.J.}},
\bauthor{\bsnm{Young}, \binits{J.-G.}},
\bauthor{\bsnm{Welles}, \binits{B.F.}}:
\batitle{A clarified typology of core-periphery structure in networks}.
\bjtitle{Science advances}
\bvolume{7}(\bissue{12}),
\bfpage{9800}
(\byear{2021})
\end{barticle}
\endbibitem

\bibitem{barbera2015critical}
\begin{barticle}
\bauthor{\bsnm{Barber{\'a}}, \binits{P.}},
\bauthor{\bsnm{Wang}, \binits{N.}},
\bauthor{\bsnm{Bonneau}, \binits{R.}},
\bauthor{\bsnm{Jost}, \binits{J.T.}},
\bauthor{\bsnm{Nagler}, \binits{J.}},
\bauthor{\bsnm{Tucker}, \binits{J.}},
\bauthor{\bsnm{Gonz{\'a}lez-Bail{\'o}n}, \binits{S.}}:
\batitle{The critical periphery in the growth of social protests}.
\bjtitle{PloS one}
\bvolume{10}(\bissue{11}),
\bfpage{0143611}
(\byear{2015})
\end{barticle}
\endbibitem

\bibitem{alvarez2005k}
\begin{botherref}
\oauthor{\bsnm{Alvarez-Hamelin}, \binits{J.I.}},
\oauthor{\bsnm{Dall'Asta}, \binits{L.}},
\oauthor{\bsnm{Barrat}, \binits{A.}},
\oauthor{\bsnm{Vespignani}, \binits{A.}}:
K-core decomposition of internet graphs: hierarchies, self-similarity and
  measurement biases.
arXiv preprint cs/0511007
(2005)
\end{botherref}
\endbibitem

\bibitem{carmi2007model}
\begin{barticle}
\bauthor{\bsnm{Carmi}, \binits{S.}},
\bauthor{\bsnm{Havlin}, \binits{S.}},
\bauthor{\bsnm{Kirkpatrick}, \binits{S.}},
\bauthor{\bsnm{Shavitt}, \binits{Y.}},
\bauthor{\bsnm{Shir}, \binits{E.}}:
\batitle{A model of internet topology using k-shell decomposition}.
\bjtitle{Proceedings of the National Academy of Sciences}
\bvolume{104}(\bissue{27}),
\bfpage{11150}--\blpage{11154}
(\byear{2007})
\end{barticle}
\endbibitem

\bibitem{kitsak2010identification}
\begin{barticle}
\bauthor{\bsnm{Kitsak}, \binits{M.}},
\bauthor{\bsnm{Gallos}, \binits{L.K.}},
\bauthor{\bsnm{Havlin}, \binits{S.}},
\bauthor{\bsnm{Liljeros}, \binits{F.}},
\bauthor{\bsnm{Muchnik}, \binits{L.}},
\bauthor{\bsnm{Stanley}, \binits{H.E.}},
\bauthor{\bsnm{Makse}, \binits{H.A.}}:
\batitle{Identification of influential spreaders in complex networks}.
\bjtitle{Nature physics}
\bvolume{6}(\bissue{11}),
\bfpage{888}--\blpage{893}
(\byear{2010})
\end{barticle}
\endbibitem

\bibitem{borgatti2000models}
\begin{barticle}
\bauthor{\bsnm{Borgatti}, \binits{S.P.}},
\bauthor{\bsnm{Everett}, \binits{M.G.}}:
\batitle{Models of core/periphery structures}.
\bjtitle{Social networks}
\bvolume{21}(\bissue{4}),
\bfpage{375}--\blpage{395}
(\byear{2000})
\end{barticle}
\endbibitem

\bibitem{tang2022patch}
\begin{bchapter}
\bauthor{\bsnm{Tang}, \binits{Y.}},
\bauthor{\bsnm{Han}, \binits{K.}},
\bauthor{\bsnm{Wang}, \binits{Y.}},
\bauthor{\bsnm{Xu}, \binits{C.}},
\bauthor{\bsnm{Guo}, \binits{J.}},
\bauthor{\bsnm{Xu}, \binits{C.}},
\bauthor{\bsnm{Tao}, \binits{D.}}:
\bctitle{Patch slimming for efficient vision transformers}.
In: \bbtitle{Proceedings of the IEEE/CVF Conference on Computer Vision and
  Pattern Recognition},
pp. \bfpage{12165}--\blpage{12174}
(\byear{2022})
\end{bchapter}
\endbibitem

\bibitem{xu2022evo}
\begin{bchapter}
\bauthor{\bsnm{Xu}, \binits{Y.}},
\bauthor{\bsnm{Zhang}, \binits{Z.}},
\bauthor{\bsnm{Zhang}, \binits{M.}},
\bauthor{\bsnm{Sheng}, \binits{K.}},
\bauthor{\bsnm{Li}, \binits{K.}},
\bauthor{\bsnm{Dong}, \binits{W.}},
\bauthor{\bsnm{Zhang}, \binits{L.}},
\bauthor{\bsnm{Xu}, \binits{C.}},
\bauthor{\bsnm{Sun}, \binits{X.}}:
\bctitle{Evo-vit: Slow-fast token evolution for dynamic vision transformer}.
In: \bbtitle{Proceedings of the AAAI Conference on Artificial Intelligence},
vol. \bseriesno{36},
pp. \bfpage{2964}--\blpage{2972}
(\byear{2022})
\end{bchapter}
\endbibitem

\bibitem{zhu2021vision}
\begin{botherref}
\oauthor{\bsnm{Zhu}, \binits{M.}},
\oauthor{\bsnm{Tang}, \binits{Y.}},
\oauthor{\bsnm{Han}, \binits{K.}}:
Vision transformer pruning.
arXiv preprint arXiv:2104.08500
(2021)
\end{botherref}
\endbibitem

\bibitem{yu2021unified}
\begin{botherref}
\oauthor{\bsnm{Yu}, \binits{H.}},
\oauthor{\bsnm{Wu}, \binits{J.}}:
A unified pruning framework for vision transformers.
arXiv preprint arXiv:2111.15127
(2021)
\end{botherref}
\endbibitem

\bibitem{chen2021chasing}
\begin{barticle}
\bauthor{\bsnm{Chen}, \binits{T.}},
\bauthor{\bsnm{Cheng}, \binits{Y.}},
\bauthor{\bsnm{Gan}, \binits{Z.}},
\bauthor{\bsnm{Yuan}, \binits{L.}},
\bauthor{\bsnm{Zhang}, \binits{L.}},
\bauthor{\bsnm{Wang}, \binits{Z.}}:
\batitle{Chasing sparsity in vision transformers: An end-to-end exploration}.
\bjtitle{Advances in Neural Information Processing Systems}
\bvolume{34},
\bfpage{19974}--\blpage{19988}
(\byear{2021})
\end{barticle}
\endbibitem

\bibitem{su2021vitas}
\begin{botherref}
\oauthor{\bsnm{Su}, \binits{X.}},
\oauthor{\bsnm{You}, \binits{S.}},
\oauthor{\bsnm{Xie}, \binits{J.}},
\oauthor{\bsnm{Zheng}, \binits{M.}},
\oauthor{\bsnm{Wang}, \binits{F.}},
\oauthor{\bsnm{Qian}, \binits{C.}},
\oauthor{\bsnm{Zhang}, \binits{C.}},
\oauthor{\bsnm{Wang}, \binits{X.}},
\oauthor{\bsnm{Xu}, \binits{C.}}:
Vitas: Vision transformer architecture search.
arXiv preprint arXiv:2106.13700
(2021)
\end{botherref}
\endbibitem

\bibitem{erdos1960evolution}
\begin{barticle}
\bauthor{\bsnm{Erdos}, \binits{P.}},
\bauthor{\bsnm{R{\'e}nyi}, \binits{A.}}, \betal:
\batitle{On the evolution of random graphs}.
\bjtitle{Publ. Math. Inst. Hung. Acad. Sci}
\bvolume{5}(\bissue{1}),
\bfpage{17}--\blpage{60}
(\byear{1960})
\end{barticle}
\endbibitem

\bibitem{watts1998collective}
\begin{barticle}
\bauthor{\bsnm{Watts}, \binits{D.J.}},
\bauthor{\bsnm{Strogatz}, \binits{S.H.}}:
\batitle{Collective dynamics of ‘small-world’networks}.
\bjtitle{nature}
\bvolume{393}(\bissue{6684}),
\bfpage{440}--\blpage{442}
(\byear{1998})
\end{barticle}
\endbibitem

\bibitem{walker1992implementing}
\begin{barticle}
\bauthor{\bsnm{Walker}, \binits{R.}}:
\batitle{Implementing discrete mathematics: combinatorics and graph theory with
  mathematica, steven skiena. pp 334. 1990. isbn 0-201-50943-1
  (addison-wesley)}.
\bjtitle{The Mathematical Gazette}
\bvolume{76}(\bissue{476}),
\bfpage{286}--\blpage{288}
(\byear{1992})
\end{barticle}
\endbibitem

\end{thebibliography}


\end{document}